\RequirePackage{fix-cm}

\documentclass[smallextended]{svjour3}       
\smartqed  
\usepackage[pdftex]{graphicx}
\usepackage{amsmath,color,cite,txfonts,tabularx,url,bm,booktabs,multirow}

\usepackage{subfig,comment}

\usepackage{caption}

\begin{document}

\title{Anime-to-Real Clothing: Cosplay Costume Generation via Image-to-Image Translation
}


\author{Koya Tango* \and Marie Katsurai* \and Hayato Maki \and Ryosuke Goto
\thanks{*Equal contribution.}
}

\authorrunning{Koya Tango et al.} 

\institute{Koya Tango* $\cdot$ Marie Katsurai*\at
            Doshisha University \\
            1-3 Tatara Miyakodani, Kyotanabe, Kyoto 610-0394, Japan \\
            Tel.: +81-774-65-7687\\
            \email{\{tango, katsurai\}@mm.doshisha.ac.jp}           %
            \and
            Hayato Maki $\cdot$ Ryosuke Goto\at
            ZOZO Technologies \\
            Aoyama Oval Building 3F, 5-52-2 Jingumae, Shibuya, Tokyo 150-0001, Japan \\
            Tel.: +81-3-3407-8811\\
            \email{hayato.maki@zozo.com} 
}

\date{Received: date / Accepted: date}

\maketitle


\begin{abstract}

Cosplay has grown from its origins at fan conventions into a billion-dollar global dress phenomenon. To facilitate imagination and reinterpretation from animated images to real garments, this paper presents an automatic costume image generation method based on image-to-image translation.
Cosplay items can be significantly diverse in their styles and shapes, and  conventional methods cannot be directly applied to the wide variation in clothing images that are the focus of this study.
To solve this problem, our method starts by collecting and preprocessing web images to prepare a cleaned, paired dataset of the anime and real domains.
Then, we present a novel architecture for generative adversarial networks (GANs) to facilitate high-quality cosplay image generation. 
Our GAN consists of several effective techniques to fill the gap between the two domains and improve both the global and local consistency of generated images. 
Experiments demonstrated that, with two types of evaluation metrics, the proposed GAN achieves better performance than existing methods.
We also showed that the images generated by the proposed method are more realistic than those generated by the conventional methods.
Our codes and pretrained model are available on the web.

\end{abstract}

\section{Introduction}
\label{intro}

Cosplay, or costume play, is a performance art in which people wear clothes to represent a specific fictional character from their favorite sources, such as \textit{manga} (Japanese comics) or \textit{anime} (cartoon animation). 
The popularity of cosplay has spanned the globe; for example, the World Cosplay Summit, which is an annual international cosplay event, attracted approximately 300,000 people from 40 countries in 2019~\cite{wcs2019about}. 
There is also an astonishing number of domestic or regional cosplay contests and conventions involving diverse creative activities.
To be a big success at these events,
it is crucial for cosplayers to wear attractive, unique, and expressive cosplay looks.
However, designing elaborate cosplay clothes requires imagination and reinterpretation from animated images to real garments.
This motivated us to devise a new system for supporting costume creation, which we call \textit{automatic costume image generation}.

\begin{figure}[t]
	\centering
	\includegraphics[clip,width=0.8\textwidth]{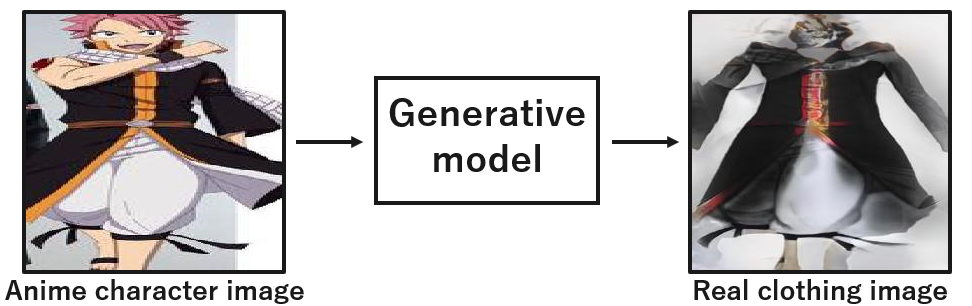}
	\caption{
		The outline of the proposed system whose aim is to generate a real clothing image for a given anime character image.
	}
	\label{fig:outline_of_the_purpose}
\end{figure}

As shown in Fig.~\ref{fig:outline_of_the_purpose}, our aim is to generate cosplay clothing item images from given anime images.
This task is a subtopic of \textit{image-to-image translation}, which learns a mapping that can convert an image from a source domain to a target domain. 
Image-to-image translation has attracted much research attention, for which several generative adversarial networks (GANs)~\cite{goodfellow2014generative} have been presented.
In the literature of fashion image translation, Yoo et al.~\cite{yoo2016pixel} trained a network that converted an image of a dressed person's clothing to a fashion product image by using multiple discriminators. 
For the same task, Kwon et al.~\cite{kwon2019coarse} introduced a coarse-to-fine scheme to reduce the visual artifacts produced by a GAN.
These conventional methods were trained using a dataset of images collected from online fashion shopping malls.
Specifically, the images were manually associated so that each pair consisted of a clothing product image and its fashion model correlate.
This dataset is high quality and less diverse because all of the product images were taken in the same position and the dressed person usually stood up straight.
However, our task of cosplay clothing synthesis presents a different hurdle to preparing such training images in that the images will not be in consistent positions.
In addition, cosplay items can be very diverse in their styles and shapes (e.g., dresses, kimonos, suits, sportswear, and school uniforms),
and the conventional methods cannot be directly applied to the wide variation in clothing images that are the focus of this study.

\begin{figure*}[t]
	\centering
		\captionsetup[subfigure]{labelformat=empty} 

	\subfloat[][Input]{
		\includegraphics[clip, width=0.85\textwidth]{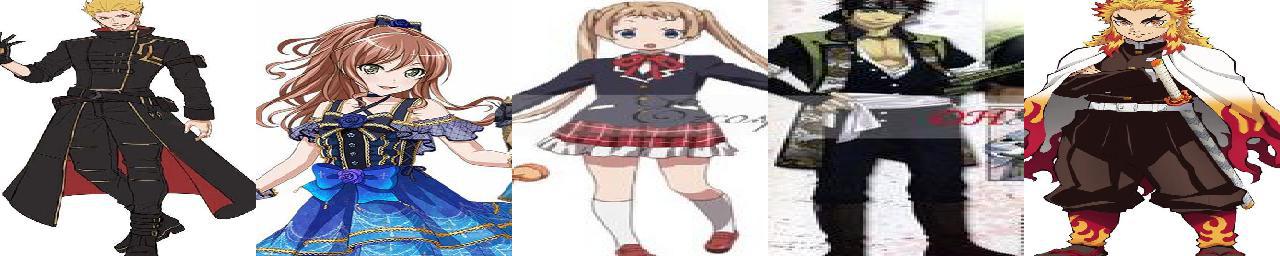}
		}\\
	\subfloat[][Ground truth]{
		\includegraphics[clip, width=0.85\textwidth]{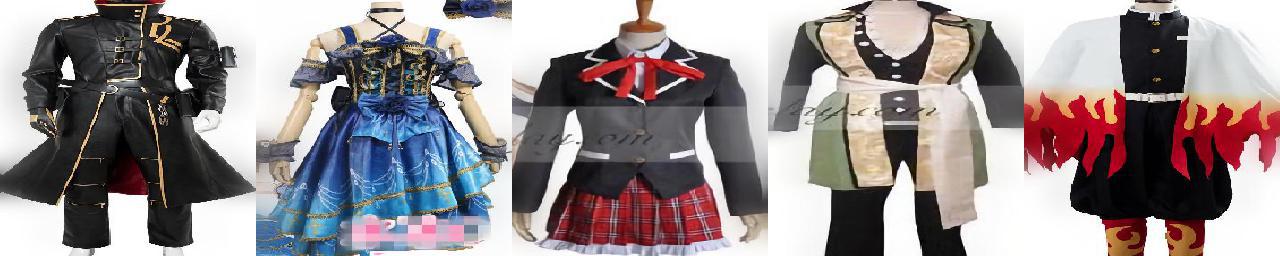}
		}\\
	\subfloat[][Ours]{
		\includegraphics[clip, width=0.85\textwidth]{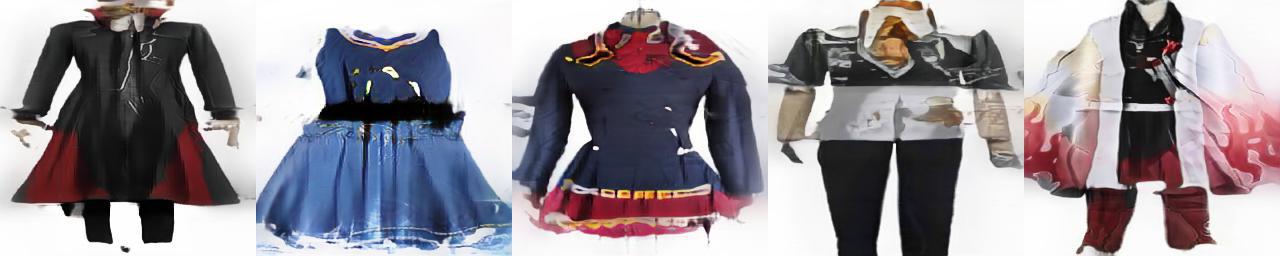}
		}
	\caption{Examples of images generated by our method. Ours represents the proposed method's translated results.}
	\label{fig:example_of_generated}
\end{figure*}

To solve this problem, this paper proposes a novel approach to cleaning a paired dataset constructed for the specific task of cosplay costume generation.
In addition, we present a novel GAN for translating anime character images to clothing images.
Our GAN architecture uses pix2pix~\cite{isola2017image} as the baseline model,
to which we introduce additional discriminators and losses that have shown  effectiveness in conventional work~\cite{wang2018high,yoo2016pixel}.
To improve the quality of image generation,
we present a novel loss, named input consistency loss, to fill the gap between the source domain (i.e., anime images) and the target domain (i.e., real clothing images). 
Results of experiments conducted using 35,633 image pairs demonstrated that our method achieved better performance than the conventional methods in terms of two quantitative evaluation measures.
Our new system will be useful not only for facilitating cosplay costume creation but also in other tasks, such as illustration-based fashion image retrieval.
Examples of the results produced by the proposed method are shown in Fig.~\ref{fig:example_of_generated}.
Our codes and the pretrained model that produced these images are available on the web.\footnote{\url{https://github.com/tan5o/anime2clothing}}

The remainder of this paper is organized as follows:
Section~\ref{sec:related-work} describes the related works.
Section~\ref{sec:section3} presents a dataset construction method for this specific task, which is one of the main novelties of this research.
Section~\ref{sec:proposed-method} presents a GAN-based anime-to-real translation method for generating cosplay costume images.
Section~\ref{sec:experimental} presents the results of the experiments conducted with a novel dataset constructed using the proposed method.
Finally, in Section~\ref{sec:conclusions}, we summarize this paper and suggest some future works. 
 
\section{Related work}
\label{sec:related-work}

GANs~\cite{goodfellow2014generative} have achieved remarkable results in a lot of computer vision tasks, including with photograph editing, super-resolution, image-to-text (or text-to-image) translation, and image-to-image translation.
A GAN-based conditional  generation approach (called a conditional GAN)~~\cite{mirza2014conditional} was proposed to assign textual tags to an input image, followed by several applications.
Examples of applying conditional GANs to the anime domain include anime colorization~\cite{ci2018user}, anime face generation~\cite{jin2017towards},
and full body generation~\cite{hamada2018full}.
GANs have also been used to synthesize realistic-looking images~\cite{cheng2020fashion}.
A typical example of a fashion synthesis application is a virtual try-on~\cite{zhu2017your, han2018viton, wu2019m2e}, which is closely related to our task.
Virtual try-on methods generally translate between real clothes within images; specifically, clothes are deformed so as to fit body shapes and poses.
However, our goal is to generate imaginary but realistic clothes given an anime image, requiring learning the mapping between different domains (i.e., real and anime).

In image-to-image translation, determining how to prepare a dataset for training models is a crucial problem.
Conventional datasets can be divided into two types: paired or unpaired image datasets. 
A paired dataset consists of an image in a source domain and its corresponding image in a target domain, while an unpaired dataset does not associate images between the source and target domains. 
One of the most famous image-to-image translation models is pix2pix~\cite{isola2017image}, which requires a paired dataset for supervised learning of the conditional GAN.
Yoo et al.~\cite{yoo2016pixel} used pix2pix for the aim of converting dressed person images to fashion product images.
They presented a domain discriminator that judges whether an input pair is actually associated, which is used in addition to the conventional real/fake discriminator.
To generate higher resolution images, pix2pix has been improved with novel
adversarial loss, multi-scale generator, and
discriminator architectures; this advanced version is called pix2pixHD\cite{wang2018high}.
Collecting images paired across source and target domains is usually time-consuming; hence, unpaired image-to-image translation frameworks have also been studied~\cite{wu2019transgaga,royer2020xgan,tang2019attention}.
Nonetheless, current methods still require that images within a single domain are consistent; for example, in the task of transferring face images across different styles, both domains' images should be taken in frontal views with a white background.
This is because a generator cannot directly learn the correspondence of shapes and colors across two different domains.
Thus, to achieve our goal, we decided to follow the paired dataset based framework
and present a useful dataset construction approach.

To improve the quality of generated images,
various GAN techniques have been proposed, such as spectral normalization~\cite{miyato2018spectral}, coarse-to-fine schemes~\cite{karras2017progressive}, and useful tips for training~\cite{salimans2016improved, heusel2017gans}. 
The design of our architecture was inspired by these successful improvements.
In particular, although our framework is based on a paired dataset,
images within a source or target domain are highly diverse in comparison to conventional problems because anime characters often wear costumes that are much more elaborate than our daily clothing.
To tackle this difficulty, we introduced a calibration strategy that aligns the position of cosplay costumes in images into the dataset construction.
In addition, our GAN architecture is equipped with a novel loss that can work to make the features of input and output images similar to each other.


\begin{figure*}[t]
	\centering
	\subfloat[][]{
		\includegraphics[clip, width=0.25\textwidth]{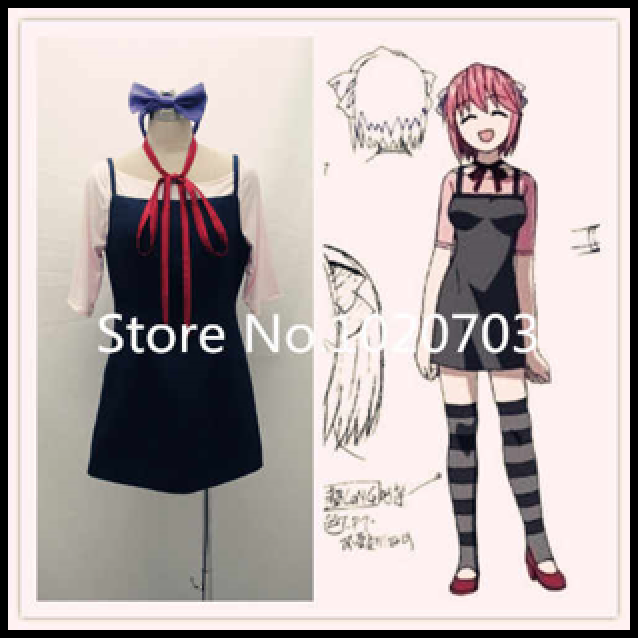}
		}
	\subfloat[][]{
		\includegraphics[clip, width=0.25\textwidth]{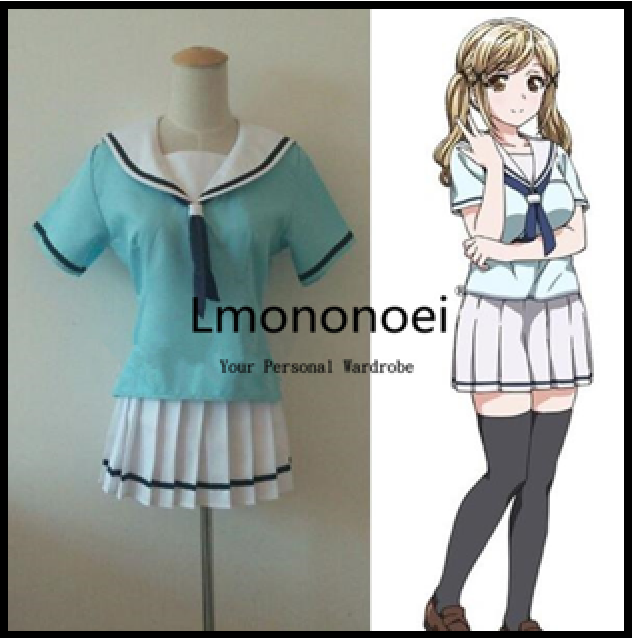}
		}
	\subfloat[][]{
		\includegraphics[clip, width=0.25\textwidth]{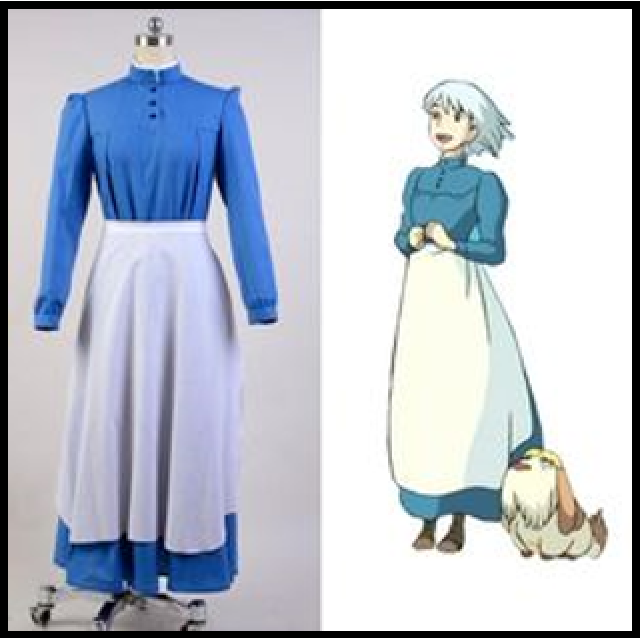}
		}
	\subfloat[][]{
		\includegraphics[clip, width=0.25\textwidth]{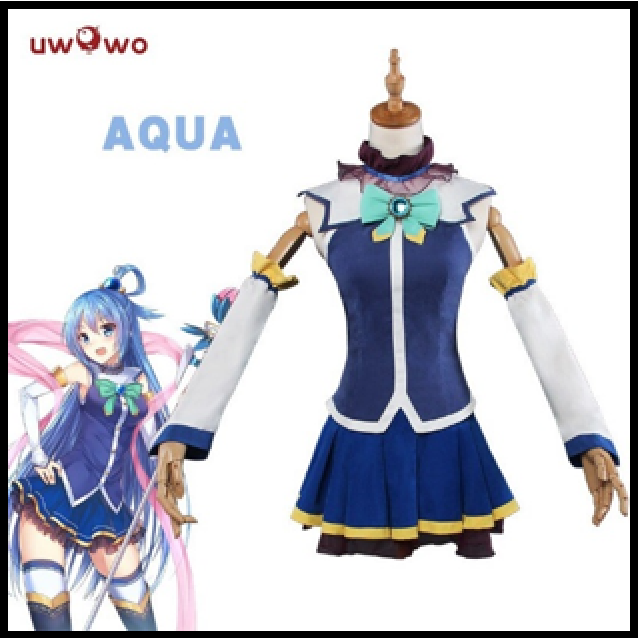}
		}
	\caption{Examples of web images we collected. }
	\label{fig:target_imgs}
\end{figure*}

\begin{figure}[t]
	\centering
	\includegraphics[clip,width=\textwidth]{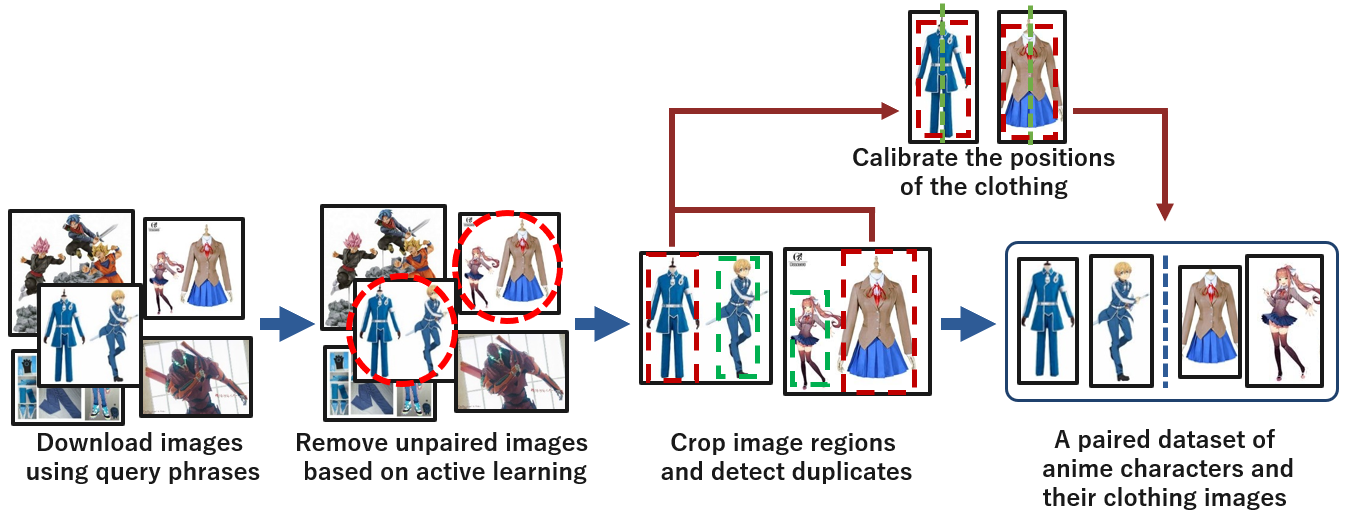}
	\caption{
		Dataset construction workflow.
	}
	\label{fig:dataset_construction_workflow}
\end{figure}

\section{Dataset construction via collecting web images}
\label{sec:section3}

Our approach requires a dataset that consists of pairs of anime characters and their corresponding cosplay clothes.
To the best of our knowledge, no public datasets comprising pairs of anime characters and clothing images is currently available on the web. 
Therefore, in this study, we focused on online cosplay shopping websites that have uploaded paired images like we wanted to use, as shown in Fig.~\ref{fig:target_imgs}.
To collect such images on the web, we first selected our query keywords by using a set of anime titles that are listed in existing datasets for anime recommendation~\cite{myanimelist, animerecodb} to create our query keywords. The query phrases used were ``\textit{cosplay costume} \textit{A} \textit{B}," in which \textit{A} and \textit{B} stood for one of the anime titles and one of 40 cosplay costume shop names, respectively. 
We downloaded images returned by the search engine,
and the total size of the images collected reached approximately 1 TB, including irrelevant or noisy images. 
To discard images that did not correspond to cosplay clothes or anime characters, we preprocessed all of the images as follows:
As shown in Fig.~\ref{fig:dataset_construction_workflow},
we first removed unpaired images using an active learning framework (see Section~\ref{ssec:extract}).
We then cropped image regions using an object detector and we also removed duplicate images (see Section~\ref{ssec:crop}).
Finally, we calibrated the positions of the clothing images for effective GAN training (see Section~\ref{ssec:calib}).

\subsection{Removing unpaired images based on active learning}
\label{ssec:extract}

Image search results are often noisy and contain images that are irrelevant to the user's query.
To efficiently detect images that were unpaired or did not contain any clothes, this study used an active learning strategy~\cite{vijayanarasimhan2011cost}, which requires manual inspection of a few samples to train the detector.
Specifically, we first manually assigned binary labels to hundreds of images so that each label indicated whether or not the corresponding image contained a character--clothes pair.
The labeled images were divided into a set of 2,760 training images and a set of 109 validation images.
We used those to fine-tune the VGG-16 model that had been pretrained on ImageNet\cite{simonyan2014very} to classify the desired pairs.
We applied the detector to additional images and manually re-labeled false classification results produced by the model, producing a larger sized training set for this detector.
Using the refined set of 3,052 training images and 215 validation images, we fine-tuned the model again because repeating these processes can improve the performance of the detector.
We applied the final detector to the whole dataset and removed images that were judged as negative.  The resulting dataset was further checked as described in the following.

\subsection{Cropping image regions and detecting duplicates}
\label{ssec:crop}
For translation model training in this study, it was desirable for an anime character and its costume to be horizontally aligned within an image, as shown in Fig.~\ref{fig:target_imgs} (a)--(c). 
However, we found that some pairs did not have this layout, such as the one shown in Fig.~\ref{fig:target_imgs} (d), and these required realignment of the images.
To crop the characters and clothing from the images,
we used single-shot detector (SSD)~\cite{liu2016ssd} for object detection.
Specifically, for the 1,059 images, we manually depicted the bounding boxes of anime characters and their clothing and used those for SSD training. The resulting model was used to detect the regions of anime characters and clothing for all images.

Because the dataset was constructed on the basis of web crawling, it contained sets of identical images. Such duplicate images in training and testing sets usually make the performance evaluation unfair.
Therefore, to reduce the number of duplicate images, we used DupFileEliminator~\cite{dupfileeliminator}, a GUI-based application that can detect highly similar images in terms of color, brightness, and contrast. We set each similarity threshold to 90\% and removed all images whose character regions were regarded as identical.
The resulting dataset consisted of 35,633 pairs of anime characters and their cosplay clothing images.

\subsection{Calibrating the positions of the clothing}
\label{ssec:calib}
\begin{figure*}[t]
	\centering
	\subfloat[][]{
		\includegraphics[clip, width=0.15\textwidth]{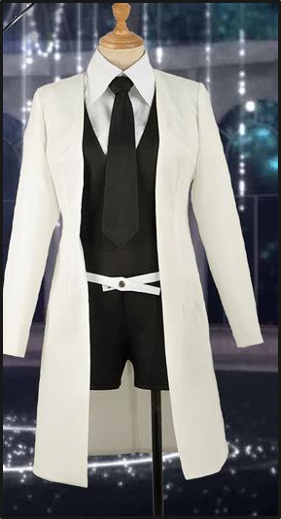}
		}
	\subfloat[][]{
		\includegraphics[clip, width=0.136\textwidth]{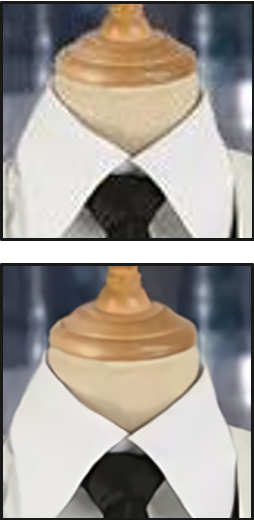}
		}
	\subfloat[][]{
		\includegraphics[clip, width=0.222\textwidth]{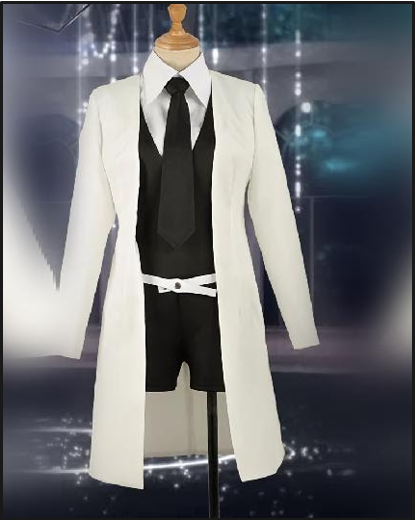}
		}
	\subfloat[][]{
		\includegraphics[clip, width=0.223\textwidth]{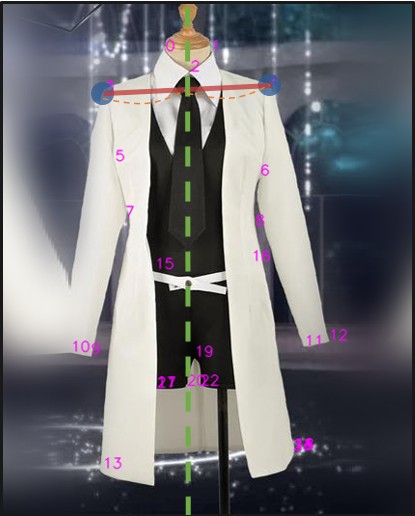}
		}
	\subfloat[][]{
		\includegraphics[clip, width=0.15\textwidth]{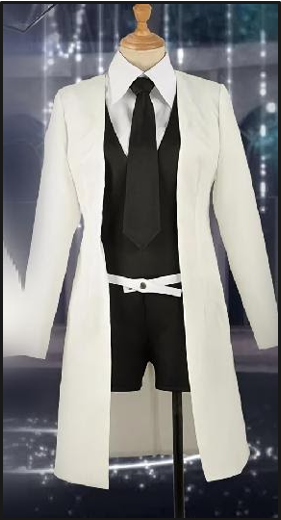}
		}
	\caption{The workflow for calibrating the position of clothing images.
	\label{chap3:fig:calibrate_image}}
\end{figure*}

Finally, we adjusted the positions of training images with the aim of reducing the difficulty of GAN training. Figure~\ref{chap3:fig:calibrate_image} shows the workflow of our image calibration.  
Given a cropped image such as the one shown in Fig.~\ref{chap3:fig:calibrate_image} (a), we applied a super-resolution method provided by \cite{waifu2xcaffe} to improve the image quality.
The upper and lower figures in Fig.~\ref{chap3:fig:calibrate_image} (b) show before and after  super-resolution, respectively.
We further applied mirror padding and Gaussian blur to both sides of the target figure as shown in Fig.~\ref{chap3:fig:calibrate_image} (c). 
On the expanded image, to detect the clothing region, we applied a fashion key point detector~\cite{vic2018fashionaikeypoint}, which is based on the Cascaded Pyramid Network~\cite{chen2018cascaded} and was pretrained using the Fashion AI Dataset\cite{zou2019fashionai}.  The center position of the clothing was calculated on the basis of the detected keypoints, as shown in Fig.~\ref{chap3:fig:calibrate_image} (d),
and we obtained images where the center positions corresponded to those of clothing regions via cropping the image width, as shown in Fig.~\ref{chap3:fig:calibrate_image} (e).

\section{Translating anime character images into clothing images}
\label{sec:proposed-method}

In this section, we propose a method for translating an anime character image into a cosplay clothing image.
The conventional pix2pix~\cite{isola2017image} is not suitable for translations that involve great changes in the shapes of target objects and would therefore have difficulty in translating an anime character image into a cosplay image. These details are first described in Section~\ref{ssec:pix2pix}. 
Our GAN architecture used pix2pix~\cite{isola2017image} as a backbone, and we introduced a novel loss with useful techniques, which are presented in Section~\ref{ssec:improve_image}.

\subsection{The pix2pix baseline}
\label{ssec:pix2pix}

The pix2pix framework consists of a generator $G$ and the domain discriminator $D_d$. 
A paired dataset of training images is denoted by $\{(x, y)\}$, where $x$ is an anime character image, and  $y$ is a cosplay clothing image. 
The aim was to make the input image closer to the target image by employing the following minimax game:

\begin{align}
    \label{eq:pix2pix}
		\underset{G}{\min} \, \underset{D_d}{\max} \mathcal{L}_{GAN_{domain}} (G,D_d) + 	\lambda\mathcal{L}_{L_{1}} (G),
\end{align}
where $\lambda$ controls the importance of the two terms, and the objective function $\mathcal{L}_{GAN_{domain}}(G,D_d)$ and the L1 loss function $\mathcal{L}_{L_{1}} (G)$ are defined as
\begin{align}
		\mathcal{L}_{GAN_{domain}}(G,D_d) = &\mathbb{E}_{(x,y)} [\log(D_d(x, y)] + \mathbb{E}_{x} [\log(1 - D_d(x, G(x)))],\\
		\mathcal{L}_{L_{1}} (G) &= \mathbb{E}_{(x,y)} [\|y - G(x)\|_1].   
\end{align}
%
%
The pix2pix framework adopts the U-Net~\cite{ronneberger2015u} as $G$
and a patch-based fully convolutional network~\cite{long2015fully} as $D_d$. The U-Net has a contracting path between the encoder and the decoder of the conventional auto encoder. This path contributes to high-quality image generation. The discriminator judges whether local image patches are real or fake, which works to improve detailed parts of the generated images.

 \begin{figure}[t]
	\centering
	\includegraphics[clip,width=0.8\textwidth]{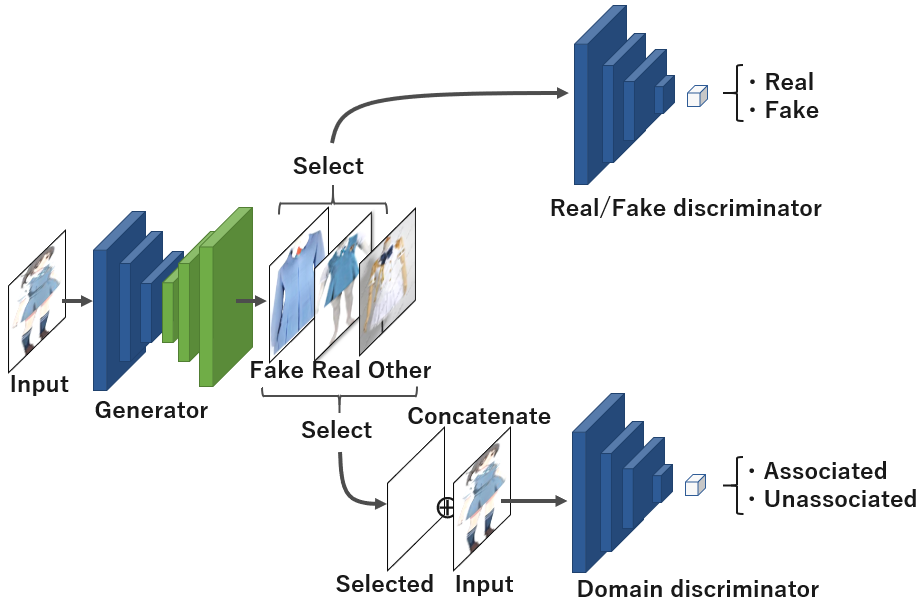}
	\caption{
		An overview of the proposed GAN architecture.
		\label{fig:overall_architecture}
	}
\end{figure}

\begin{figure}[ht]
\centering
\begin{tabular}{c}
		\includegraphics[clip, width=0.75\textwidth]{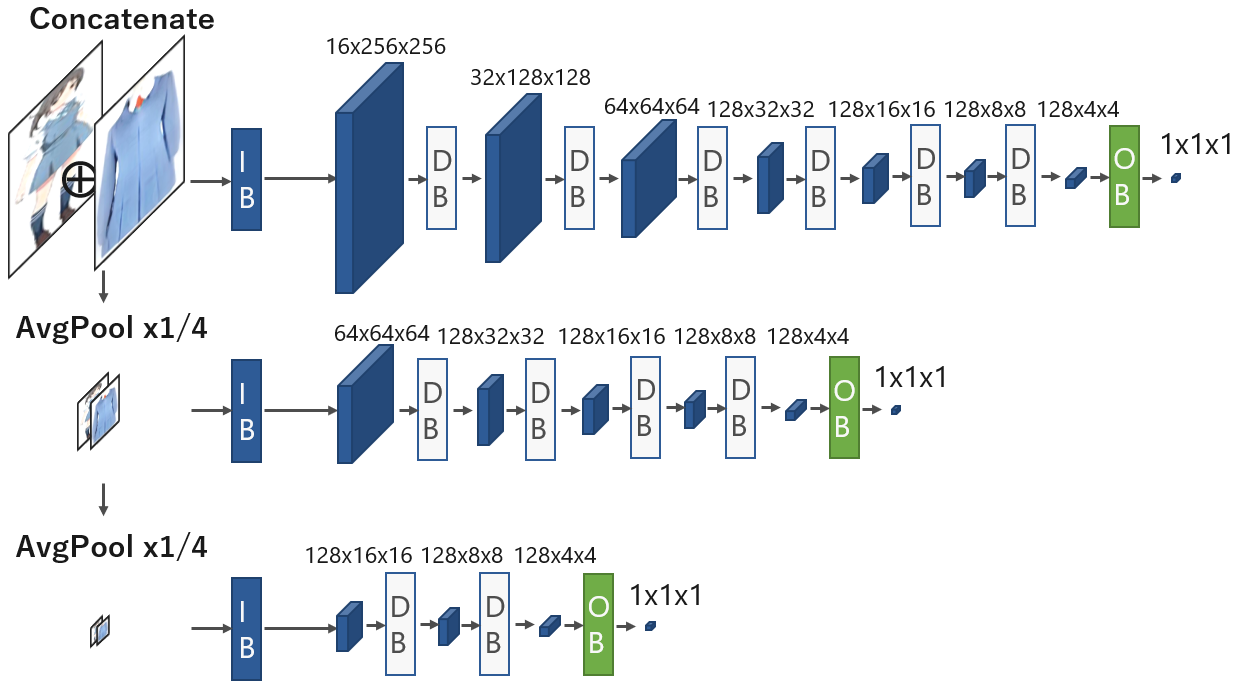}
		\\
		(a) The multi-scale domain discriminator $D_d$.\\ \\
		\includegraphics[clip, width=0.75\textwidth]{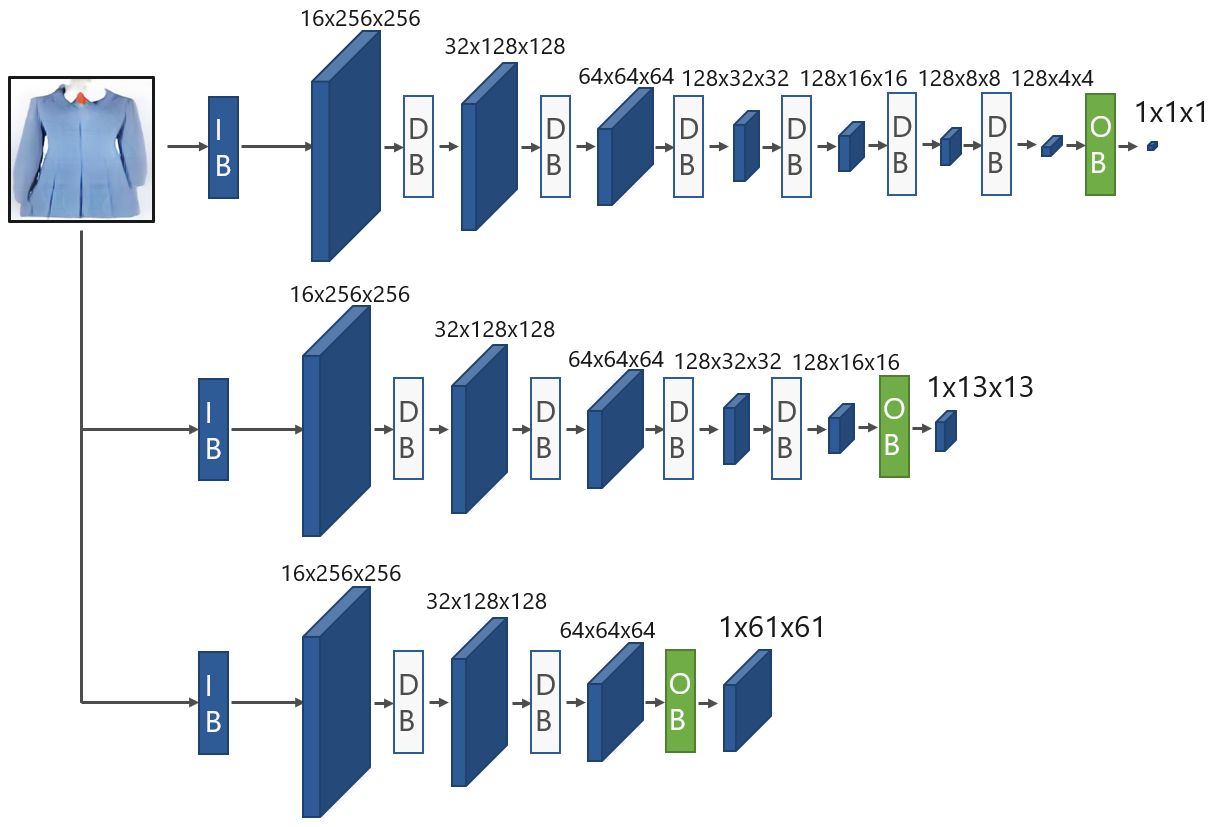}
		\\
		(b) The real/fake multi-scale patch discriminator $D_r$.\\ \\
		\includegraphics[clip, width=0.75\textwidth]{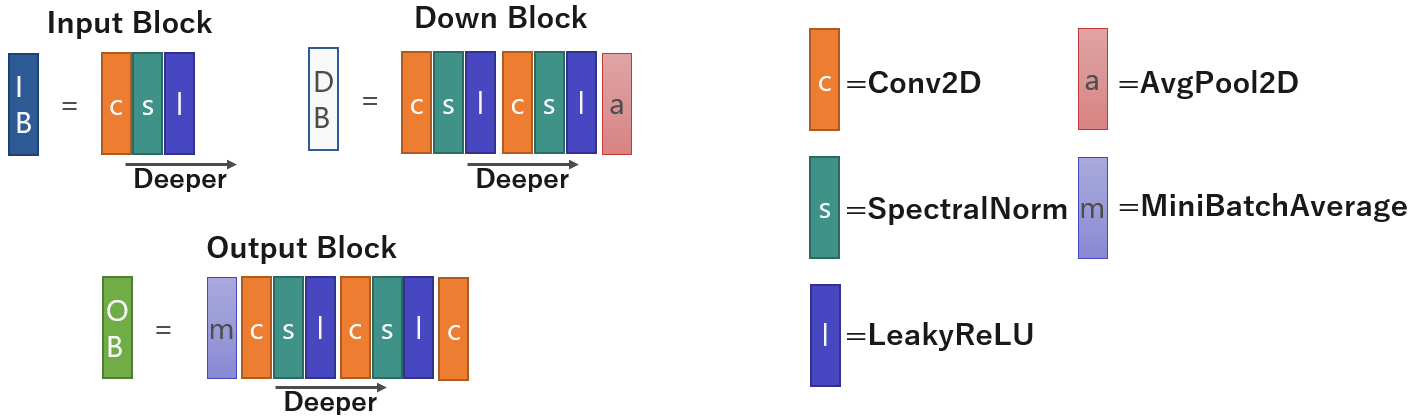}
		\\
		(c) The Structure of discriminator blocks in (a) and (b).
	\end{tabular}
	\caption{
		Structures of our model's discriminators.
		\label{fig:discriminator_architecture}
	}
\end{figure}

As seen in Eq.~\eqref{eq:pix2pix}, pix2pix combines adversarial learning and the L1 loss. 
Minimizing the L1 loss aims to allow for learning the correspondence of shapes and colors between two image domains whose visual compositions are closely related (e.g., building images and their semantic segmentation images~\cite{cordts2016cityscapes}).
However, in the case of translation from the anime character image to the cosplay clothing image, the shapes of objects in these images do not correspond with each other and so cannot be captured well by the L1 loss. Furthermore, adversarial learning that focuses on local image patches lacks the ability to improve the global quality of generated images.

\subsection{The proposed GAN architecture}
\label{ssec:improve_image}

To solve the problems with applying the conventional pix2pix framework in our task,
we present a novel GAN architecture that has an improved domain discriminator with additional functions: the real/fake discriminator, the feature matching loss, and the input consistency loss.
This subsection describes the details of each technique.
Figure~\ref{fig:overall_architecture} shows the overall architecture of our GANs, and Fig.~\ref{fig:discriminator_architecture} shows the architectures of our discriminators.

\subsubsection{Improved domain discriminator}
For effective anime-to-real translation,
two techniques were used to improve the domain discriminator $D_d$.
First, to stabilize high-quality image generation from coarse to fine levels,
we extended the discriminator to a multi-scale architecture~\cite{wang2018high}, as shown in Fig.~\ref{fig:discriminator_architecture} (a).
Specifically, using three discriminators, $D_1$, $D_2$, and $D_3$, which have an identical network structure but operate at different image scales,
the objective is modified as a multi-task learning of
%
\begin{align}
\underset{G}{\min}\, \underset{D_d}{\max} &\sum_{k=1}^3 \mathcal{L}_{{GAN}_{domain}} (G,D_{d_k}), \\
D_d &= \{D_{d_1}, D_{d_2}, D_{d_3}\}.\notag
\end{align}
Solving the above problem guides the generator $G$ to generate globally consistent images, keeping fine details.
We applied spectral normalization~\cite{miyato2018spectral} to each discriminator to satisfy the Lipschitz constraint.

Second, we designed modified supervision of the domain discriminator.
The original pix2pix's domain discriminator determines whether an input image and the generated image are paired; specifically, given an input image, it regards its real ground truth image and any synthesized image as True and False, respectively.
However, this supervision is too weak to capture the relationship between the anime and real domains.
Inspired by related work~\cite{yoo2016pixel}, we added new False judgments: for each input image, we provided False labels to other images when they were real clothes but did not correspond to the input image's truly associated image. This new associated/unassociated supervision can facilitate the transfer of an anime character image into a more realistic and relevant clothing image.

\subsubsection{Real/fake discriminator}
Using the domain discriminator alone, it is difficult to generate images with substantial shape changes, such as a translation from anime characters to clothing images. This is because the domain discriminator only determines whether the two input images are associated or unassociated and does not specialize in the quality of the generated image. 
To improve the image quality, we needed to use a discriminator that checks the quality of the generated image, which is called a real/fake discriminator~\cite{yoo2016pixel}.
Hereafter, the real/fake discriminator is denoted by $D_r$. 

Similar to the improved domain discriminator, this real/fake discriminator should also contribute to high-quality image generation. Here, to maintain local consistency, we propose a multi-scale patch discriminator similar to the structure reported in \cite{shocher2018ingan}.
As shown in Fig.~\ref{fig:discriminator_architecture} (b), our multi-scale patch discriminator outputs three different patches: $1\times 1$, $13\times 13$, and $61\times 61$ feature maps.
These different discriminators are denoted by $D_{r_1}$, $D_{r_2}$, and $D_{r_3}$, respectively.
Using multiple patch sizes allowed for capturing both fine-grained details and coarse structures in the image. 
The GAN generates images via the following minimax game:
\begin{align}
\underset{G}{\min}\, \underset{D_r}{\max} &\sum_{k=1}^3 \mathcal{L}_{GAN_{real/fake}} (G,D_{r_k}), \\
D_r &= \{D_{r_1}, D_{r_2}, D_{r_3}\},\notag
\end{align}
\noindent
where the objective function $\mathcal{L}_{GAN_{real/fake}}(G,D_r)$ is given by

\begin{align}
    \mathbb{E}_{y} [\log(D_r(y))] + \mathbb{E}_{x} [\log(1 - D_r(G(x)))].
\end{align}

\noindent
We also applied spectral normalization to each discriminator.

\subsubsection{Feature matching loss}
 The feature matching loss was originally presented by \cite{wang2018high} with the aim of generating an image closer to a corresponding real one.
 For a given pair of real and synthesized images, it is computed as the L1 loss between outputs of the intermediate layer of a discriminator.
We designated $D^{(i)}$ as the $i$-th layer of the discriminator. 
In the proposed method, the feature matching loss based on the domain discriminator and the feature matching loss based on the real/fake discriminator are defined as follows:
\begin{align} 
	\mathcal{L}_{FM_{domain}} (G,D_d) = \mathbb{E}_{(x,y)} \sum_{i=1}^{T} {N_i} \bigg[ \bigg\|{D_d^{(i)}(x, y) - D_d^{(i)}\bigl(x, G(x) \big) }\bigg\| _{1} \bigg], \label{eq:feature_match1}\\
	\mathcal{L}_{FM_{real/fake}} (G,D_r) = \mathbb{E}_{(x,y)} \sum_{i=1}^{T} {N_i} \bigg[\bigg\|D_r^{(i)}(y) - D_r^{(i)}(G(x))\bigg\|_{1}\bigg], \label{eq:feature_match2}
\end{align} 
where $T$ is the total number of layers, and $N_i$ denotes the number of elements in each layer~\cite{wang2018high}.

 \begin{figure}[t]
	\centering
	\includegraphics[clip,width=0.8\textwidth]{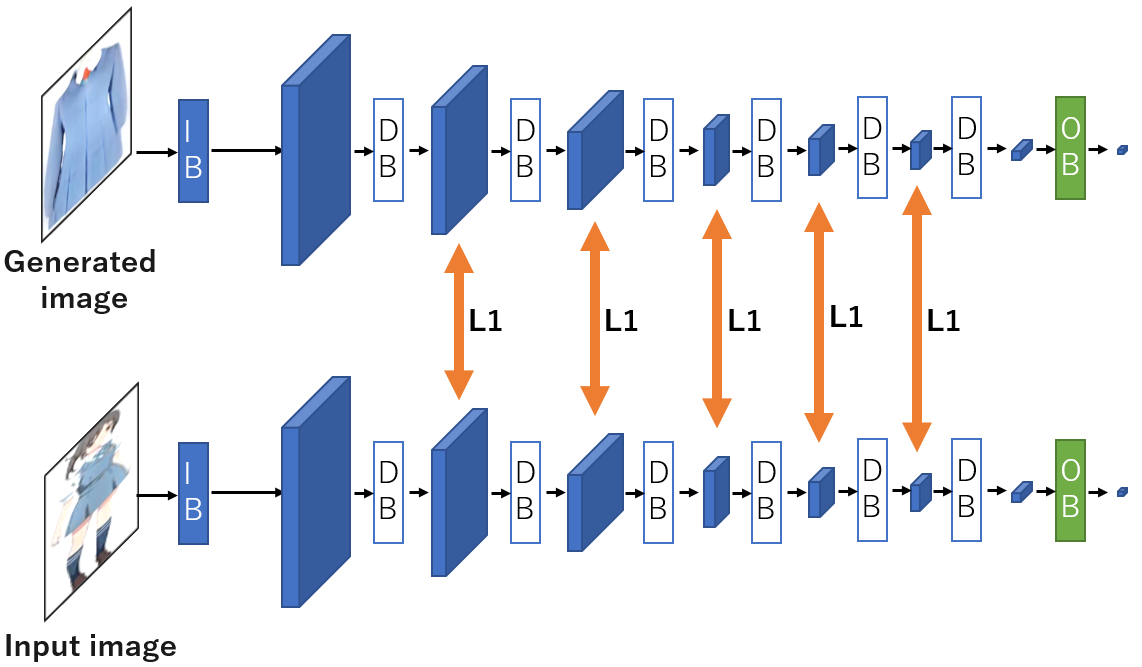}
	\caption{
		An overview of calculating the input consistency loss, which is based on the L1 loss of the matrix extracted from the intermediate layers of the discriminator.
		\label{fig:input_consistency}
	}
\end{figure}

\subsubsection{Input consistency loss}
In our task, the generator had to generate a real clothing image that retained the detailed shape and color information of the original anime character image. To maintain consistency between an input image and its corresponding output image, we propose a method of minimizing the differences between them called the input consistency loss. Figure~\ref{fig:input_consistency} shows our calculation overview. As shown, we computed the L1 loss over all intermediate layers as follows:

\ 
\begin{align}
    \label{eq:input_loss}
	\begin{split}
	\mathcal{L}_{input_{real/fake}} (G,D_r) = \mathbb{E}_{x} \sum_{i=1}^{T} {N_i} \bigg[\bigg\|D_r^{(i)}(x) - D_r^{(i)}(G(x))\bigg\|_{1}\bigg].
	\end{split}
\end{align}
\

\begin{figure}[t]
	\centering
    \captionsetup[subfigure]{labelformat=empty} 
    \subfloat[][]{
		\includegraphics[clip, width=0.9\textwidth]{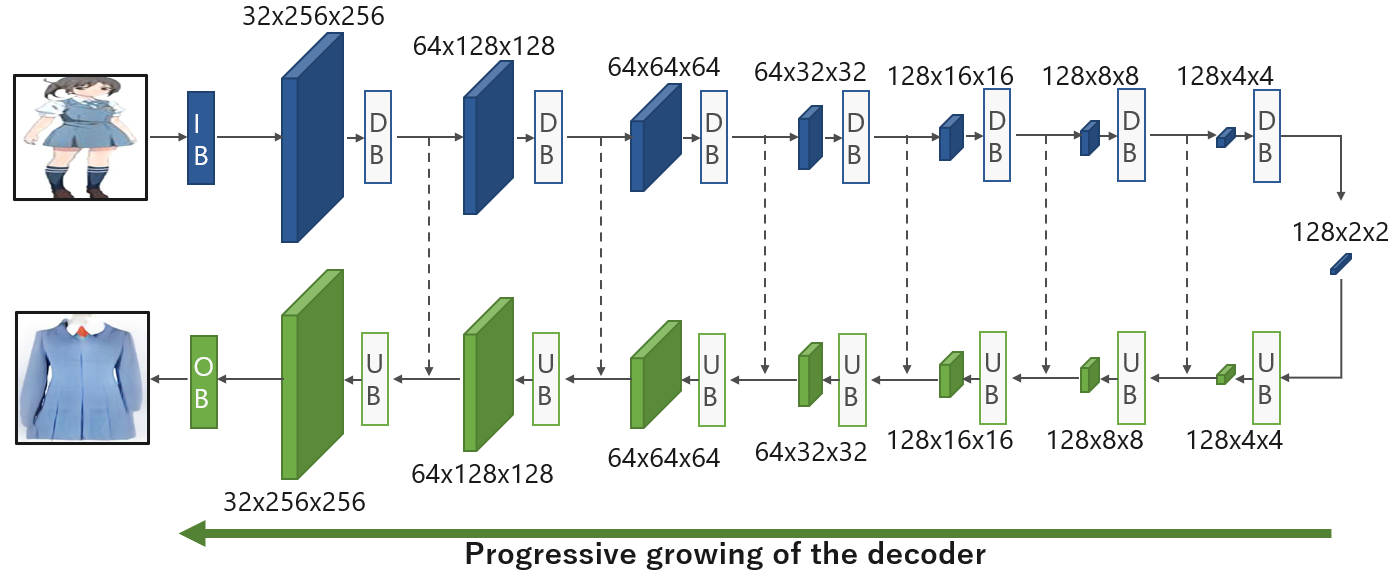}
		}\\
	\subfloat[][]{
		\includegraphics[clip, width=0.8\textwidth]{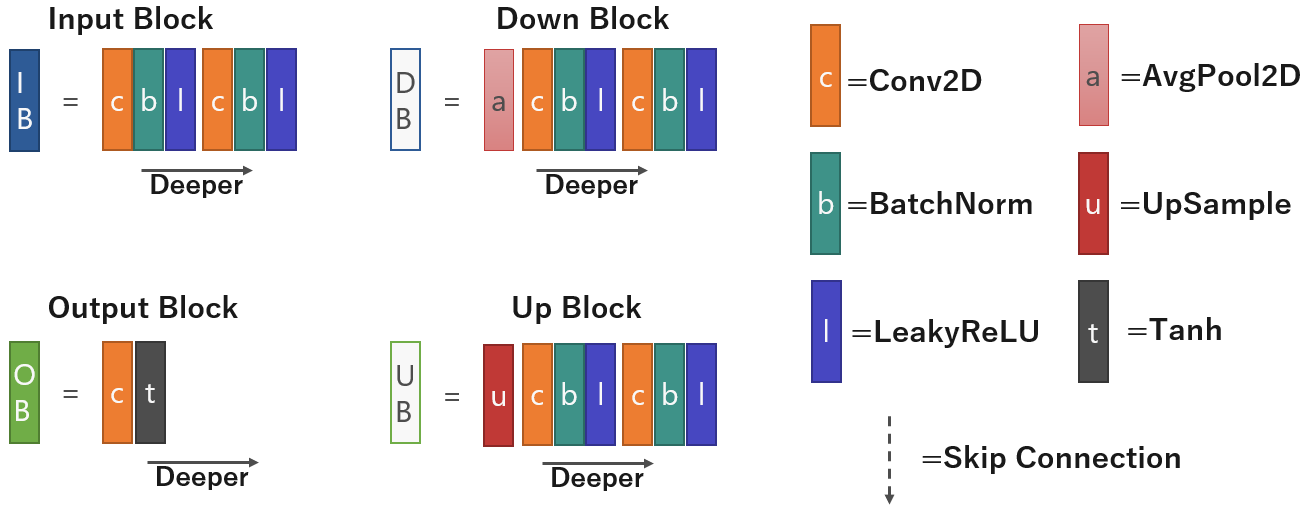}
		}
	\caption{
		The structure of our U-Net generator.
		\label{fig:generator_architecture}
	}
\end{figure}

\subsubsection{Training}
Finally, using the U-Net as the generator whose structure is shown in Fig.~\ref{fig:generator_architecture},
our full objective was calculated based on the combination of the GAN objectives, the feature
matching loss, and the input consistency loss, as follows.

\ 
\begin{align}
\label{eq:full_objective}
\begin{split}
\underset{G}{min} \, \biggl(\underset{D_d,\:{D}_r}{max} \Bigl( \sum_{k=1}^3  \mathcal{L}_{GAN_{domain}} (G,D_{d_k}) + \sum_{k=1}^3 \mathcal{L}_{GAN_{real/fake}} (G,D_{r_k}) \Bigl) \\+ \sum_{k=1}^3  \mathcal{L}_{FM_{domain}} (G,D_{d_k}) + \sum_{k=1}^3 \mathcal{L}_{input_{real/fake}} (G,D_{r_k}) + 	\lambda\mathcal{L}_{L_{1}} (G) \biggl). \\
\end{split}
\end{align}
%
Training GANs based on high-resolution images often results in the gradient problem and takes a lot of computation time.
To alleviate this problem, we used a coarse-to-fine scheme for training the  GAN~\cite{karras2017progressive}.  
This scheme progressively grows both the generator and discriminator: starting from low-resolution images, we added new
layers that introduced higher resolution details as the training progressed.
The effectiveness of this coarse-to-fine scheme was also investigated in the experiments.


\

\section{Experimental evaluation}
\label{sec:experimental}

\subsection{Implementation details}

We divided the dataset into 32,608 training images and 3,025 testing images.
All the networks were trained using the Adam optimizer~\cite{kingma2014adam} with the initial learning rate $\alpha=0.0002$ and momentum parameters $\beta_1=0.5$, and $\beta_2=0.99$ on two NVIDIA RTX 2080Ti GPUs. After 70 epochs, the learning rate $\alpha$ was linearly decayed over the next 30 epochs. The initial weights were sampled from the normal distribution with the zero mean and a standard deviation of 0.02. To avoid overfitting, we applied the following data augmentation techniques~\cite{shorten2019survey} to the input anime images: three-degree random rotations, a random crop, and a hue and saturation jitter of 0.05. In addition, we applied random horizontal flips to the clothing images. For 
the adversarial loss of GANs, we used the squared difference as proposed in LSGANs~\cite{mao2017least}. Throughout all the experiments, we used $\lambda = 10$ in Eq.~\eqref{eq:full_objective} and $N_1$ = 5, $N_{2,3,4}$ = 1.5, and $N_5$ = 1 in  Eqs.~\eqref{eq:feature_match1} and \eqref{eq:feature_match2}.



\subsection{Evaluation metrics}
The performance of a paired image-to-image translation system should be evaluated from two points of view: the image-based similarities and the distribution-based similarities. First, because every source image of a system has a corresponding target image, the generated and the target images should be similar to each other. Second, images generated in the same domain can reasonably be assumed to be sampled from an identical probability distribution, and hence, the probability distributions that the sets of the generated and the target images follow should be similar. The first and second points were measured using the Learned Perceptual Image Patch Similarity (LPIPS)~\cite{zhang2018unreasonable} and the Fr\'echet Inception Distance (FID)~\cite{heusel2017gans}, respectively.
\\

\noindent
{\bf Fr\'echet Inception Distance (FID)}~\cite{heusel2017gans}:
Assuming that images in the same domain are sampled from an identical probability distribution, the FID measures the divergence between two probability distributions that the sets of generated and target images follow, respectively. Specifically, the FID computes the Wasserstein-2 distance between distributions approximated with the Gaussian,
which were estimated using the vectors extracted from the intermediate layer of the inception networks~\cite{szegedy2016rethinking} when inputting the generated/target images.
The FID score should get lower as an image synthesis system gets better. 
\\

\noindent
{\bf Learned Perceptual Image Patch Similarity (LPIPS)}~\cite{zhang2018unreasonable} :
The LPIPS quantifies how similar two images are using the feature vectors extracted from the intermediate layer of the AlexNet~\cite{krizhevsky2012imagenet}. A lower LPIPS value means that the generated image is perceptually similar to the actual image.\footnote{Please note that this metric name is similarity, but the smaller it is, the more similar it is.}



\subsection{Comparisons among different configurations with the proposed method}

\begin{figure}

\begin{minipage}{0.9\textwidth}
	\makeatletter
    \centering
	\def\@captype{table}
	\makeatother
	\caption{
		Comparison of training configuration. The proposed method corresponds to configuration (j), showing improvement from the baseline in FID and LPIPS.
		\label{tab:training_configuration}
	}
    
	\centering
	\begin{tabular}{@{}llllll@{}}
	\toprule
	\textbf{Configuration} & \textbf{\footnotesize FID} & \textbf{LPIPS} \\ \midrule
	{ (a) Baseline} & 52.05 & 0.5957 \\
	{ (b) +Coarse-to-fine scheme}  & 148.52 & 0.6131 \\
	{ (c) +Calibrate dataset}  & 153.37 & 0.5970 \\
	{ (d) +Real/fake discriminator } & 92.98 & \textbf{0.5672}  \\
	{ (e) +Unassociated pair } & 109.60 & 0.5674 \\
	{ (f) +Spectral normalization } & 40.80 & 0.5923\\
	{ (g) +Multi-scale discriminator } & 53.47 & 0.5933 \\
	{ (h) +Multi-scale patch discriminator } & 38.96 & 0.6023 \\
	{ (i) +Feature match }  & 35.11 & 0.5940 \\
	{ (j) +Input consistency loss } & \textbf{30.38} & 0.5786 \\ \midrule
	{ (k) Ground truth} & 0.00   & 0.0000 \\ \bottomrule
	\end{tabular}
\end{minipage}

\bigskip

\begin{minipage}{\textwidth}
	\captionsetup[subfigure]{labelformat=empty} 

	\subfloat[][Input]{
		\includegraphics[clip, width=0.152\columnwidth]{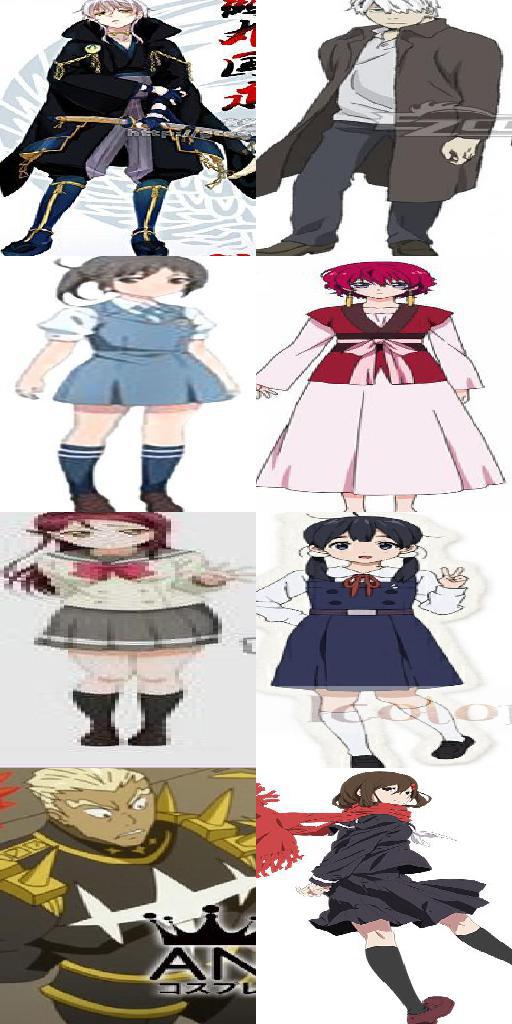}
		}
	\addtocounter{subfigure}{-1}
	\captionsetup[subfigure]{labelformat=parens} 
	\subfloat[][]{
		\includegraphics[clip, width=0.152\columnwidth]{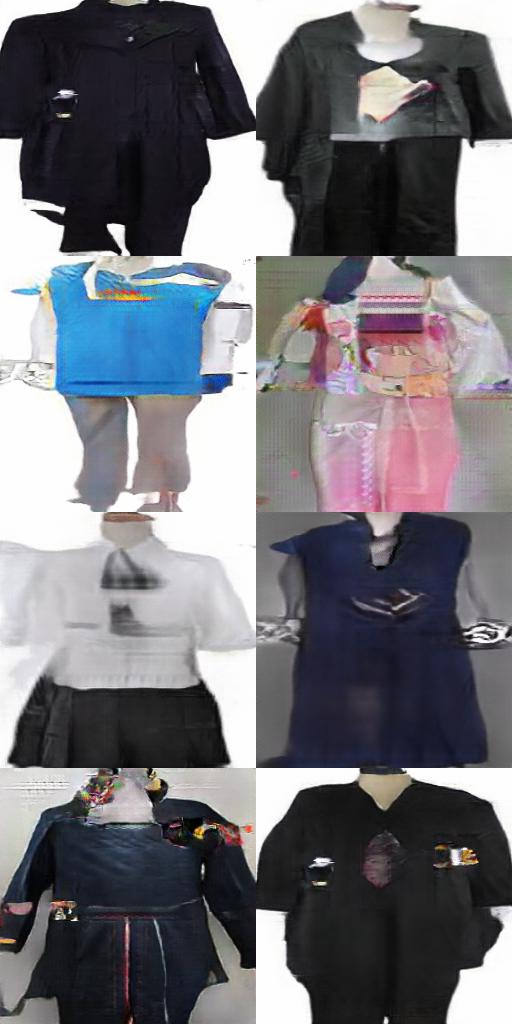}
		}
	\subfloat[][]{
		\includegraphics[clip, width=0.152\columnwidth]{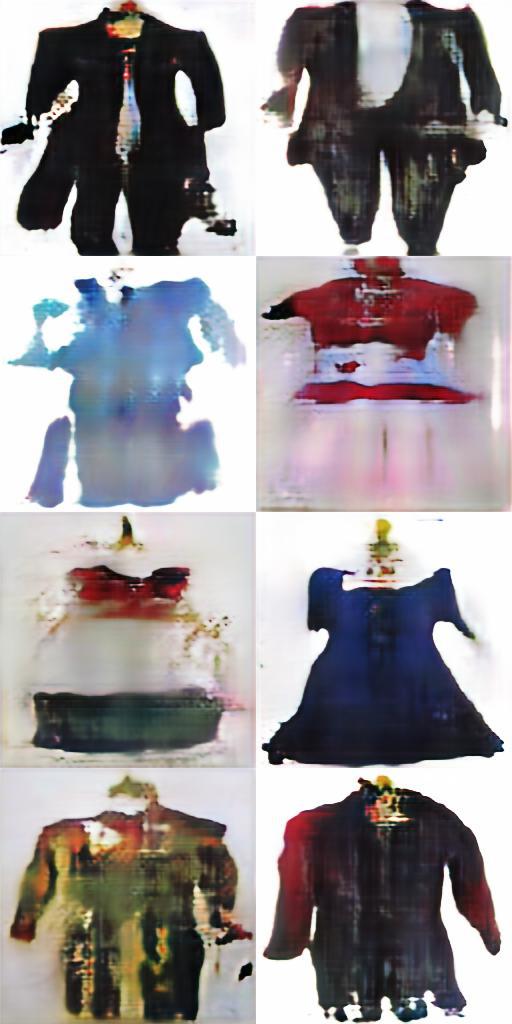}
		}
	\subfloat[][]{
		\includegraphics[clip, width=0.152\columnwidth]{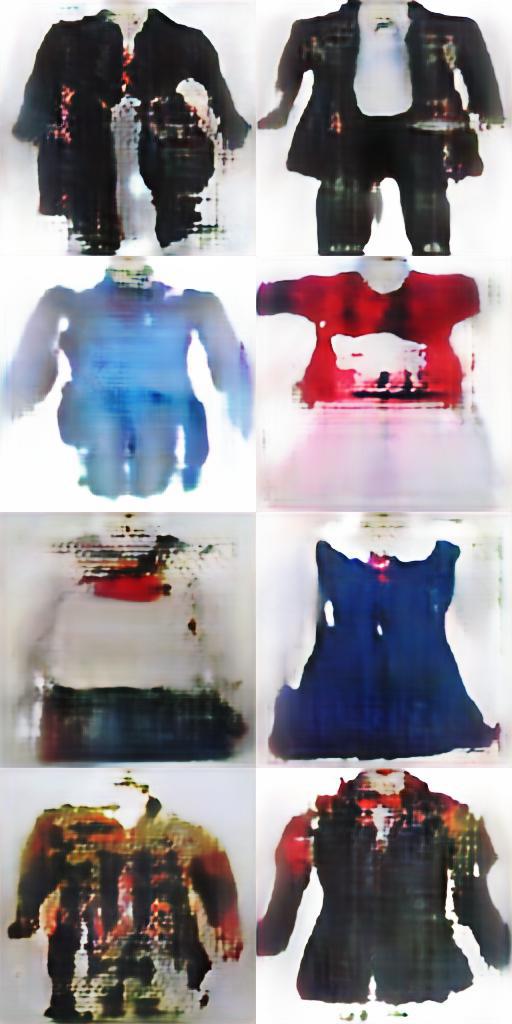}
		}
	\subfloat[][]{
		\includegraphics[clip, width=0.152\columnwidth]{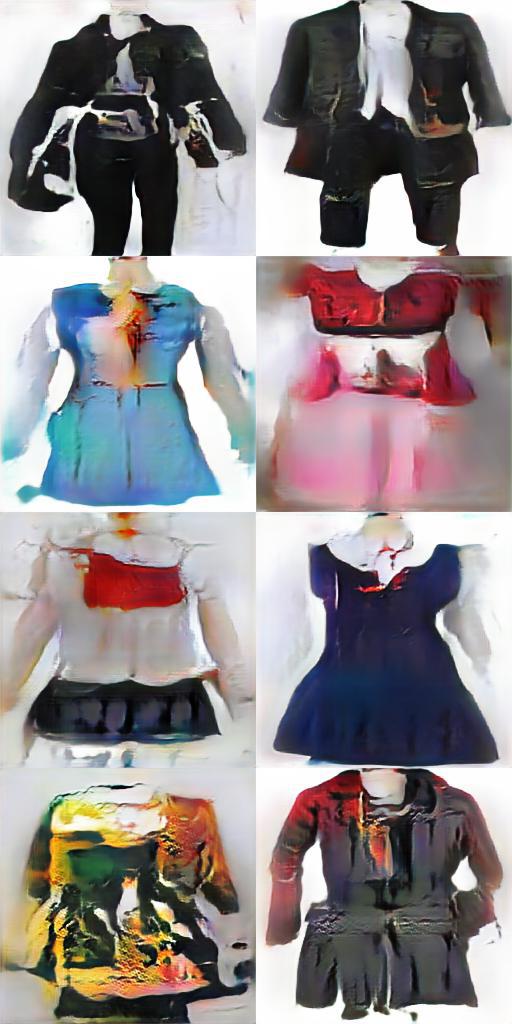}
		}
	\subfloat[][]{
		\includegraphics[clip, width=0.152\columnwidth]{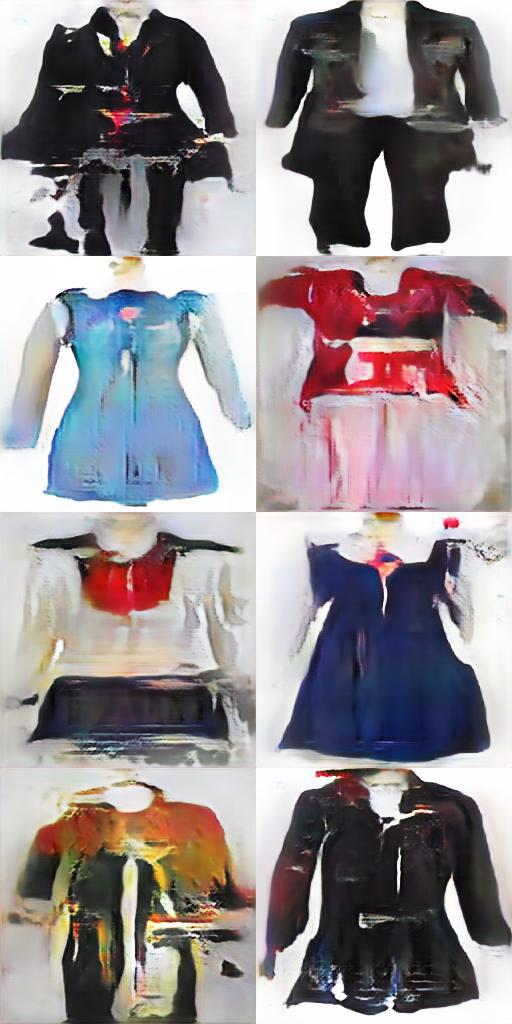}
		}\\
	\subfloat[][]{
		\includegraphics[clip, width=0.152\columnwidth]{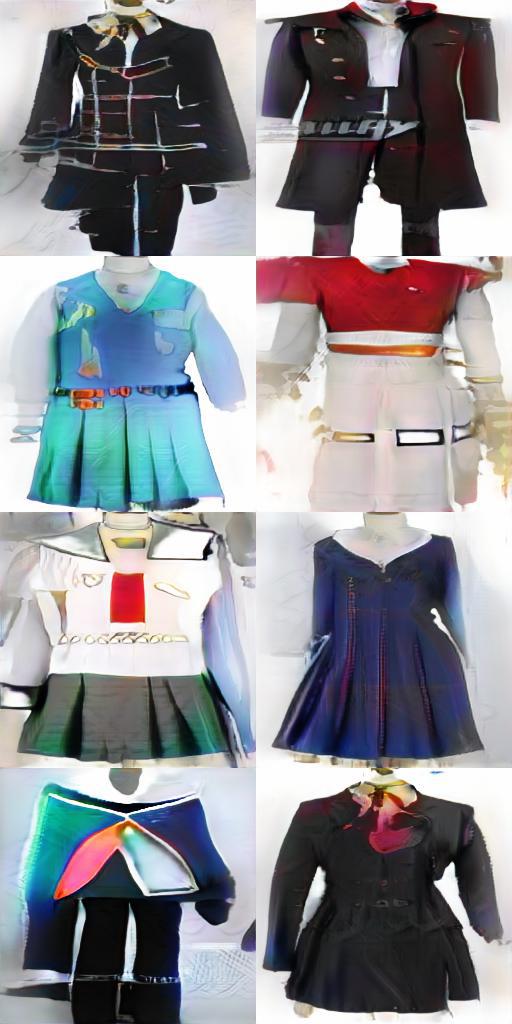}
		}
	\subfloat[][]{
		\includegraphics[clip, width=0.152\columnwidth]{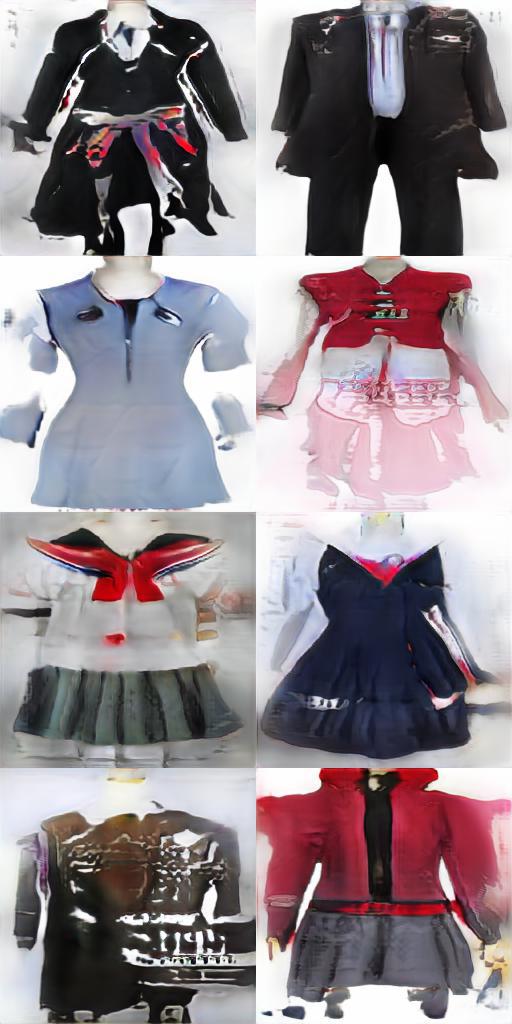}
		}
	\subfloat[][]{
		\includegraphics[clip, width=0.152\columnwidth]{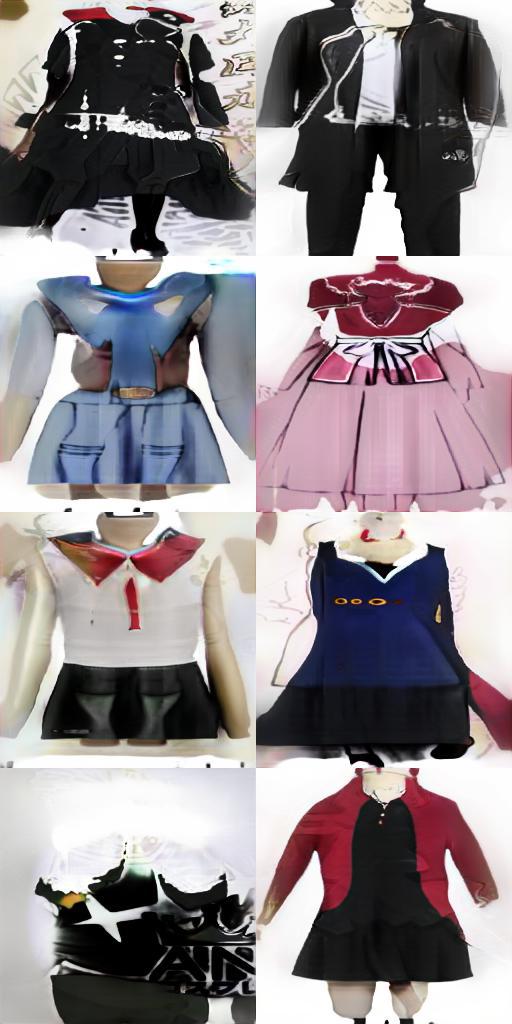}
		}
	\subfloat[][]{
		\includegraphics[clip, width=0.152\columnwidth]{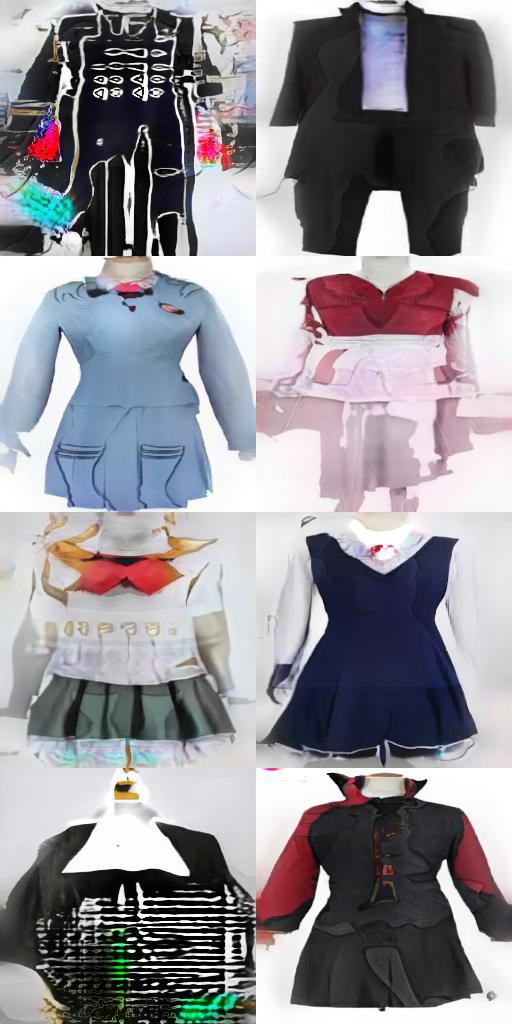}
		}
	\subfloat[][]{
		\includegraphics[clip, width=0.152\columnwidth]{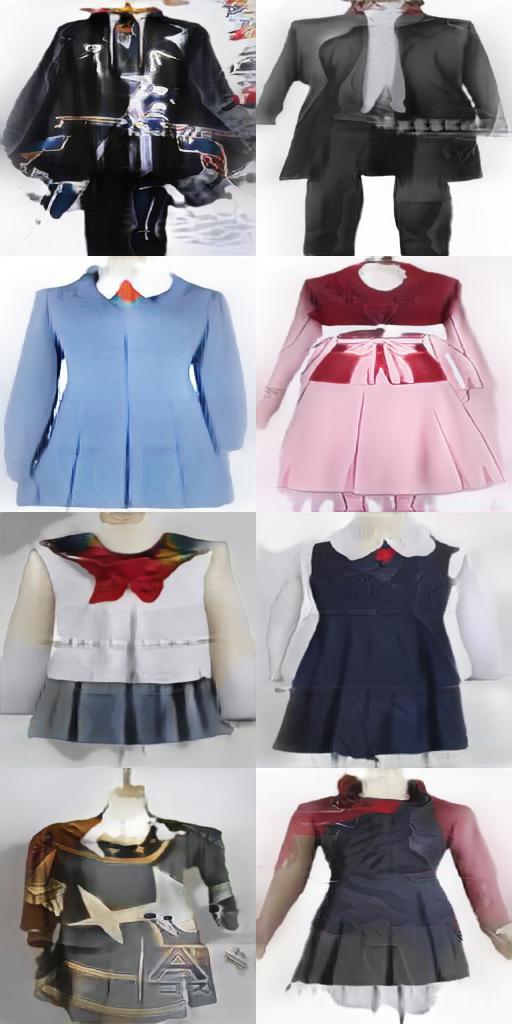}
		}
	\subfloat[][GT]{
		\includegraphics[clip, width=0.152\columnwidth]{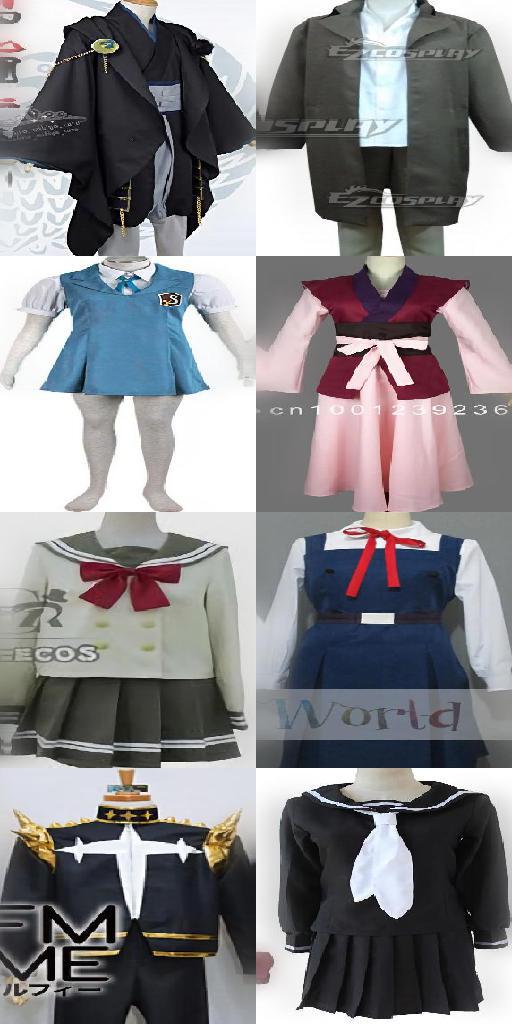}
		}
	\caption{Results of image generation with different configurations (a)--(i), which correspond to the rows of Table~\ref{tab:training_configuration}. GT means the ground truth.}
	\label{fig:comparison_results}

\end{minipage}
\end{figure}

To evaluate the effectiveness of each technique described in Section~\ref{ssec:improve_image}, we experimented with several configurations of the proposed method.
The quantitative evaluation results are shown in Table~\ref{tab:training_configuration}. 
The baseline (a) used the conventional pix2pix~\cite{isola2017image}, as explained in Section~\ref{ssec:pix2pix}. In (b), we added the coarse-to-fine scheme, where we changed the size of the discriminator's feature maps from $70\times 70$ to $1\times 1$.\footnote{Our training at (b) (i.e., adding the coarse-to-fine scheme only) became unstable, and hence, we changed the discriminator's instance normalization~\cite{ulyanov2017improved} to the equalized learning rate~\cite{karras2017progressive} to stabilize the training of (b). We canceled this setting in (c).} 
In (c), we calibrated the positions of all clothing images to validate the effectiveness of our dataset construction approach. 
In (d), we added the real/fake discriminator~\cite{yoo2016pixel}. This discriminator significantly improved the FID as compared with (c). It enabled the generator to output images similar to the ground truth images. 
In (e), we added unassociated pairs with False labels to the domain discriminator's training.
In (f), we applied spectral normalization to each discriminator, which made the training stable and significantly improved the FID.
In (g), we used the multi-scale discriminator as $D_d$. 
In (h), we used the multi-scale patch discriminator as $D_r$, leading the FID's improvement.
In (i), we added the feature matching losses $\mathcal{L}_{FM_{domain}} (G,D_d)$ and $\mathcal{L}_{FM_{real/fake}} (G,D_r)$. These contributed to generating images that captured global and local consistency, improving both FID and LPIPS, as compared with (h).
Finally, in (j), we added the input consistency loss, completing the proposed method.
Our method achieved the best FID score as well as a better LPIPS score than the baseline.
Figure~\ref{fig:comparison_results} shows the clothing image generation results for eight testing images as provided by each configuration. 
We observed in Fig.~\ref{fig:comparison_results} that adding each technique to the architecture had effectively improved the image quality.
In particular, as shown in Fig.~\ref{fig:comparison_results} (h) and Fig.~\ref{fig:comparison_results} (j), the multi-scale patch discriminator $D_r$ and the input consistency loss contributed to describing the textures in detail.

\subsection{Comparisons with conventional methods}

\begin{figure}

\begin{minipage}{\textwidth}
    \centering
	\makeatletter
	\def\@captype{table}
	\makeatother
	\caption{
	Comparisons with conventional methods.
	\label{tab:comparison_exist_method}
	}
	\centering
	\begin{tabular}{@{}lllllll@{}}
	\toprule
	\textbf{Method} & \textbf{\footnotesize FID} & \textbf{\footnotesize LPIPS} \\ \midrule
	PLDT~\cite{yoo2016pixel}   & 248.47 & 0.8220 \\
	pix2pix~\cite{isola2017image}   & 55.44 & 0.5820 \\
	pix2pixHD~\cite{wang2018high} & 197.59 & 0.6857 \\
	\textbf{Ours} & \textbf{30.38} & \textbf{0.5786} \\ \bottomrule
	\end{tabular}
\end{minipage}

\bigskip

\begin{minipage}{\textwidth}
	\centering
	\captionsetup[subfigure]{position=top, labelformat=empty, justification=raggedright, font=footnotesize,} 

	\subfloat[Input]{
		\includegraphics[clip, width=0.16\columnwidth]{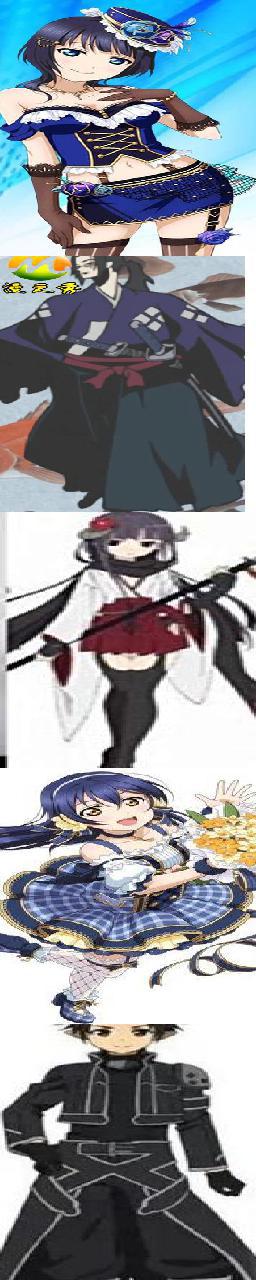}
		}
	\subfloat[GT]{
		\includegraphics[clip, width=0.16\columnwidth]{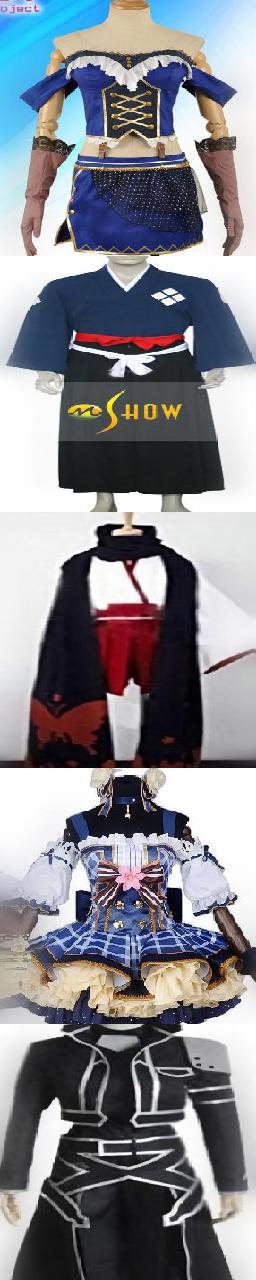}
		}
	\hspace{3pt}
	\subfloat[PLDT]{
		\includegraphics[clip, width=0.16\columnwidth]{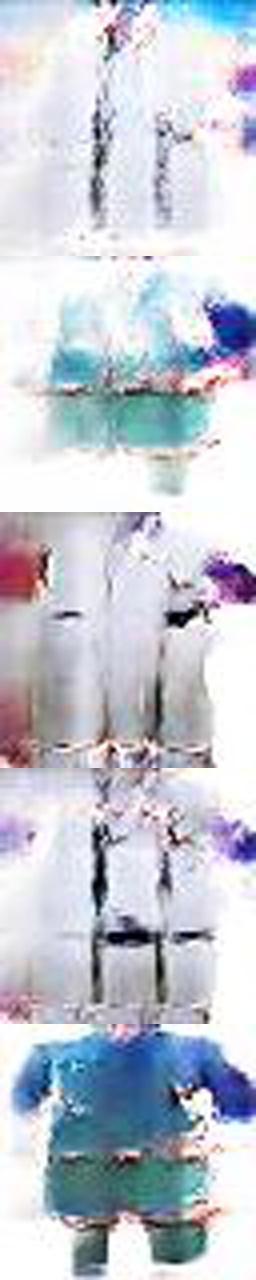}
		}
	\hspace{-8pt}
	\subfloat[pix2pix]{
		\includegraphics[clip, width=0.16\columnwidth]{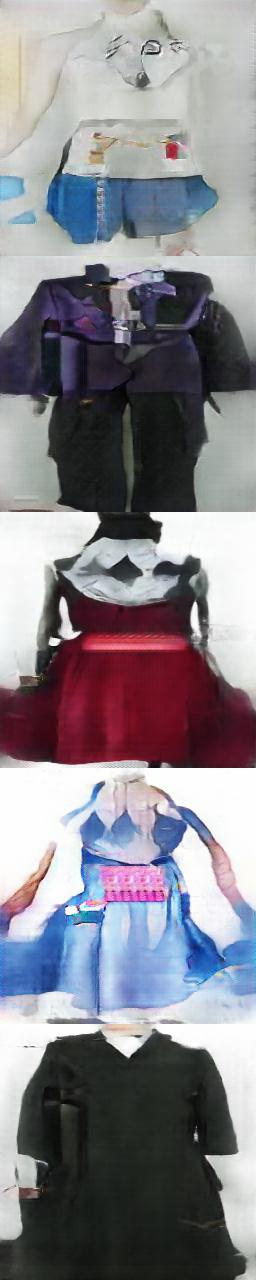}
		}
	\hspace{-8pt}
	\subfloat[pix2pixHD]{
		\includegraphics[clip, width=0.16\columnwidth]{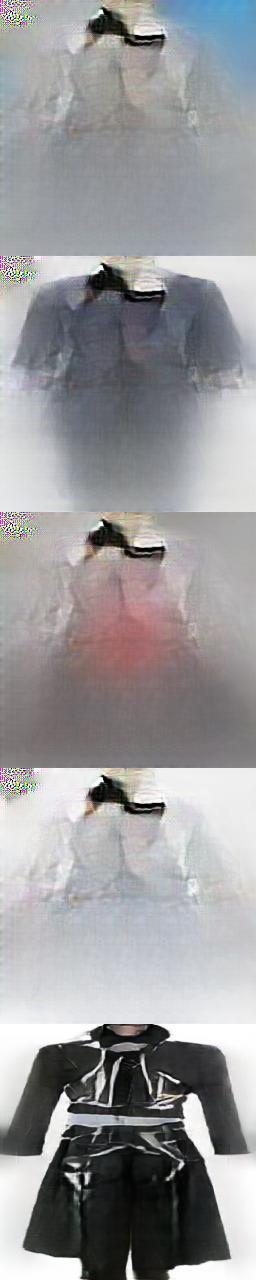}
		}
	\hspace{-8pt}
	\subfloat[\textbf{ours}]{
		\includegraphics[clip, width=0.16\columnwidth]{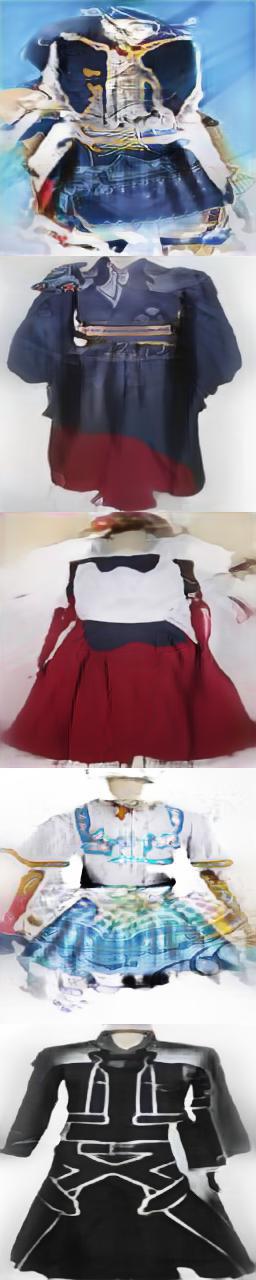}
		}
	\caption{Comparisons with the conventional image-to-image translation methods. GT means the ground truth.}
	\label{fig:comparison_other}

\end{minipage}
\end{figure}

Using the dataset constructed with our approach as outlined in Section~\ref{sec:section3},
we compared our method with the following three conventional methods: pixel-level domain transfer (PLDT)~\cite{yoo2016pixel}, pix2pix~\cite{isola2017image}, and pix2pixHD~\cite{wang2018high}. The PLDT was trained with a batch size of 128. Because the resolution of the PLDT's outputs was originally $64\times 64$, it was enlarged to $256\times 256$ using the bicubic interpolation based on $4\times 4$ pixels.  The pix2pix was trained with the batch size of 1, using its official implementation. The pix2pixHD was trained with a batch size of 16 and a resolution of $256\times 256$, using its official implementation.
All GANs were trained using the same data augmentation as ours.
Table~\ref{tab:comparison_exist_method} shows  a comparison with other image-to-image translation methods. 
As shown, our method achieved the best performance in terms of both FID and LPIPS.
Pix2pix does not have an L1 loss, and its training was unstable.
Although pix2pixHD is an advanced version of pix2pix, it suffered from overfitting and failed to generate high-quality images.
Note that the difference between the baseline in Table~\ref{tab:training_configuration} (a) and pix2pix in Table~\ref{tab:comparison_exist_method} is whether or not the proposed dataset calibration approach was used.
Figure~\ref{fig:comparison_other} shows the clothing image generation results obtained by the proposed method and the conventional methods for five testing images.
Characters in these testing images wore diverse and complex costumes.
We can see that our GAN effectively produced images that are realistic and have fine-grained detail as compared with the conventional methods.
Our pretrained model, available on the web, thus demonstrated that it can produce such clothing images for any anime character image.

\section{Conclusion and Further Work}
\label{sec:conclusions}

This paper proposed a novel method for translating anime character images into  clothing images for facilitating cosplay costume creation.
We first described an approach to constructing a clean, paired dataset for our task.
Then, we presented a novel GAN architecture equipped with several techniques to fill the gap between anime and real and to improve the generated image quality.
The experiments conducted using our dataset demonstrated that the proposed GAN  achieved the best performance among several existing methods in terms of both FID and LPIPS.
We also showed that the images generated by the proposed method were more realistic than those generated by the conventional methods, using five testing images.

Our method still has room for improvement.
Although our dataset calibration was effective for GAN training, there are some possible outliers in the collected noisy web images.
The proposed input consistency loss is currently calculated based on the L1 loss between an input anime character and its synthesized clothes image, assuming the body proportions of the characters are constant to some extent over all training images. 
If the face or head of a character in the input image is significantly larger than its clothes (e.g., Hello Kitty\footnote{\url{https://www.sanrio.co.jp/character/hellokitty/}}), the current generator might fail to output a relevant clothing image: the clothes' colors and shapes must be affected by the face or head.
Our future work includes the development of a more sophisticated approach that considers the pixel-level correspondence between the input and output.
In addition, we will investigate how to transform an input character's body shape to be consistent throughout the dataset.


%
%

\bibliographystyle{spmpsci}      
\bibliography{reference}

\end{document}